\documentclass[11pt,a4paper]{article}
\usepackage[hyperref]{emnlp-ijcnlp-2019}
\usepackage{times}
\usepackage{graphicx}
\usepackage{latexsym}
\usepackage{amsmath}

\usepackage{url}

\aclfinalcopy % Uncomment this line for the final submission

\title{Teacher-Student Framework Enhanced Multi-domain Dialogue Generation}

\author{Shuke Peng$^1$, Xinjing Huang$^1$, Zehao Lin$^1$, Feng Ji$^2$\footnote{Corresponding Author}, Haiqing Chen$^2$, Yin Zhang$^1$ \\
  $^1$Digital Media Computing \& Design Lab, Zhejiang University, China \\
  {\tt \{pengsk, huangxinjing, georgelin, zhangyin98\}@zju.edu.cn} \\
  $^2$DAMO Academy, Alibaba Group, Hangzhou, China
  \\
  {\tt \{zhongxiu.jf, haiqing.chenhq\}@alibaba-inc.com} 
}

\date{}

\begin{document}
\maketitle
\begin{abstract}
Dialogue systems dealing with multi-domain tasks are highly required. How to record the state remains a key problem in a task-oriented dialogue system. Normally we use human-defined features as dialogue states and apply a state tracker to extract these features. However, the performance of such a system is limited by the error propagation of a state tracker. In this paper, we propose a dialogue generation model that needs no external state trackers and still benefits from human labeled semantic data. By using a teacher-student framework, several teacher models are firstly trained in their individual domains, learn dialogue policies from labeled states. And then the learned knowledge and experience are merged and transferred to a universal student model, which takes raw utterance as its input. Experiments show that the dialogue system trained under our framework outperforms the one uses a belief tracker.  \end{abstract}

\section{Introduction}
Spoken Dialogue Systems(SDS) are widely used as assistants to help users in processing daily affairs. Tasks often vary from searching for a restaurant to booking several flight tickets. The demand for finishing tasks in diverse situations requires the SDS to have the ability to handle different domains of the dialogue.

To build a successful SDS, state representation is an essential part of an end-to-end dialogue system. A general used method is to use a human-defined state representation, where the state records necessary information such as indispensable slot values the system needs. The dialogue policy then makes actions associating with the state. To generate such representation in the real dialogue situation, a belief state tracker is usually adopted to recognize the ontology from the user's text.

Another optional way is to use a hidden state representation. The text is compressed into hidden vectors from the raw utterance. The model summaries the context and dialogue acts are making from the hidden states. In this setting, the dialogue system is pure an end to end model with only text as the input.

Both two methods above have their restraints. The model with human-defined states as input is bounded by an attached state tracking model. 
The errors accumulated in the states tracking processing, especially in a multi-domain situation where the space of ontology is large. And the abandoning of human-defined states leads to the poor ability of the model. Besides a model with the latent state is always hard to understand and debug for humans. 

In this paper, we introduce a universal dialogue generation system dealing with multi-domain dialogues. Rather than using an external tracker to recognize the ontology, our model straightly generates hidden states from raw text. And to make a proper response and benefit from well labeled semantic on training data, we brought a teacher-student framework. In the framework multiple teachers are applied in every required domain to learn the dispersed dialogue knowledge and labeled extra semantic information. Then we extract and merge the well learned knowledge and policy methods from individual teacher models into a universal student model. The framework ensures the student studies the well-learned responses during conversations. Our model takes full advantage of the labeled data and is not bounded by the performance of an outside belief tracker.

The main contributions of our work can be briefly summarized in two folds:
\begin{itemize}
    \item We use a multi-teacher single-student framework to gather the extra knowledge from individual domains to one universal model in dialogue systems.
    
    \item We built a multi-domain dialogue system that takes no human-defined states as input and can still benefit from semantic labeling while training.
\end{itemize}

\section{Related Works}
\subsection{Multi-domain Dialogue System}
 The architectures of a dialogue system usually consist of following key components: A Spoken Language Understanding(SLU) that understands the users' intents, a dialogue manager that captures the dialogue states and makes decisions for the response, a Natural Language Generator(NLG) that generates human-readable text responses. 
 Task-oriented dialogue systems in special domains process problems according to domain-specific ontology. The increase in the scale of domains depends on the improved ability of every component above.
 
 A large number of works have been done with how to construct a dialogue system in multi-domain. Such as dynamic dialogue processing for users as a daily assistant\cite{pakucs2003towards}, the scalable action spaces for a dialogue system to share knowledge between domains\cite{dzikovska2003integrating}. As the neural network is widely used in dialogue systems, ideas appeared to handle the multi-domain dialogue with a deep network. For example, \citet{wen2016multi} brings the method that improves the dialogue system's ability in one domain by pre-train the neural network from another domain. And \citet{ultes2017pydial} developed a multi-domain dialogue system toolkit with the implementations of all dialogue system modules such as a DQN-based dialogue policy.
 
 \subsection{Dialogue State}
 The dialogue states clarify the current step the dialogue processing locates in. The dialogue states can be regarded as a Partially Observable Markov Decision Processes(POMDPs)\cite{thomson2010bayesian}. A general way to represent such processes is to apply human-defined features and consider the multi-hot embedding vectors as the states. These features often contain slots that must be filled in the task, and domain tags if there is more than one domain. States generated by this kind of embedding method are often well explainable. 
 
 To apply this method into practical application, we need an external state tracker to recognize correct features from user utterance. Many works have been done on this problem such as a rule-based state tracker \cite{sun2014generalized} or a Neural Belief Tracker(NBT)\cite{mrkvsic2016neural}. There are also works focusing on state trackers that track user intent and slot values in multiple domains. \cite{mrkvsic2015multi,rastogi2017scalable,goel2018flexible,nouri2018toward}
 
 Another method in dialogue state representation is to use the hidden state vector generated directly from the raw text. A Hierarchical Recurrent Encoder-Decoder(HRED)\cite{sordoni2015hierarchical,serban2016building,serban2017hierarchical} dialogue system summarizes the history dialogues by utterance vectors without handcrafted features in an open domain. And an Attention with Intention(AWI)\cite{yao2015attention} architecture did the work in a similar way. 
 These models take only the raw text as their inputs and outputs, and need no human defined state information while training, thus get rid of a state tracker component.
 
 \subsection{The Teacher-Student framework}
 The teacher-student framework illustrates a teacher model guiding the training step to generate a better student by its internal knowledge. 
 The idea of the teacher-student framework in deep learning is brought in knowledge distillation by \citet{hinton2015distilling}, where the knowledge is extracted from a large teacher model to a small one or assembled from several models into one student. The earliest application of knowledge distillation is mainly in computer vision. Recent works show that knowledge distillation based teacher-student method works well in a language model\cite{kim2016sequence}. 
 \citet{fan2018learning} extends the teacher-student framework that the duty of the teacher is no longer simply transferring the knowledge but deciding what kind of data to learn, in what space of hyper-parameters, and how well the student can reach.
 And \citet{tan2019multilingual} proposed a multi-teachers-single-student framework that combines more than one individual model to a multilingual model in the language translation task.
 
\section{Dialogue Generation Systems}
Our multi-domain dialogue generation system can be illustrated as three parts: A multi-domain hierarchical dialogue generation model, serving as the main model and the student model to learn external knowledge. Several individual dialogue models, take the roles of teacher models to guide a student. And the guiding step that transfers the knowledge from individual models to the universal one.

The problem of multi-turn dialogues can be considered as a sequence to sequence mapping problem. At the time $t$, the user inputs an utterance $u_t$, the dialogue finds the most proper response $r_t$ according to $u_t$ and the history context $(u_0, ..., u_{t-1}, r_0, ..., r_{t-1})$. That is,  maximum $p(r_t|u_{0\sim  t},r_{0\sim t-1})$. By introducing POMDPs, the history dialogues can be summarized as the state $s_t$. Usually, the state is generated from the all user utterance and the history responses from the system $s_t=f(u_0, ..., u_t, r_0, ..., r_{t-1})$. In a reinforcement learning setting of dialogue problem, we use actions to represent what the system should respond. The dialogue policy makes actions from the states $a_t=\pi(s_t)$. And the response is generated from the NLG module by the corresponding action $r_t=g(a_t)$, or by both the action and the user's utterance in an attention mechanism enhanced dialogue model, $r_t=g(a_t, u_t)$. 

In our model, the state $s_t$ for the teacher model is directly defined from human labeled semantic in the utterance, and the student model generates the $s_t$ itself from all passed user utterance. The detail of the two models will be discussed in the rest of this section.

\begin{figure}[!t]
    \centering
    \includegraphics[width=0.5\linewidth]{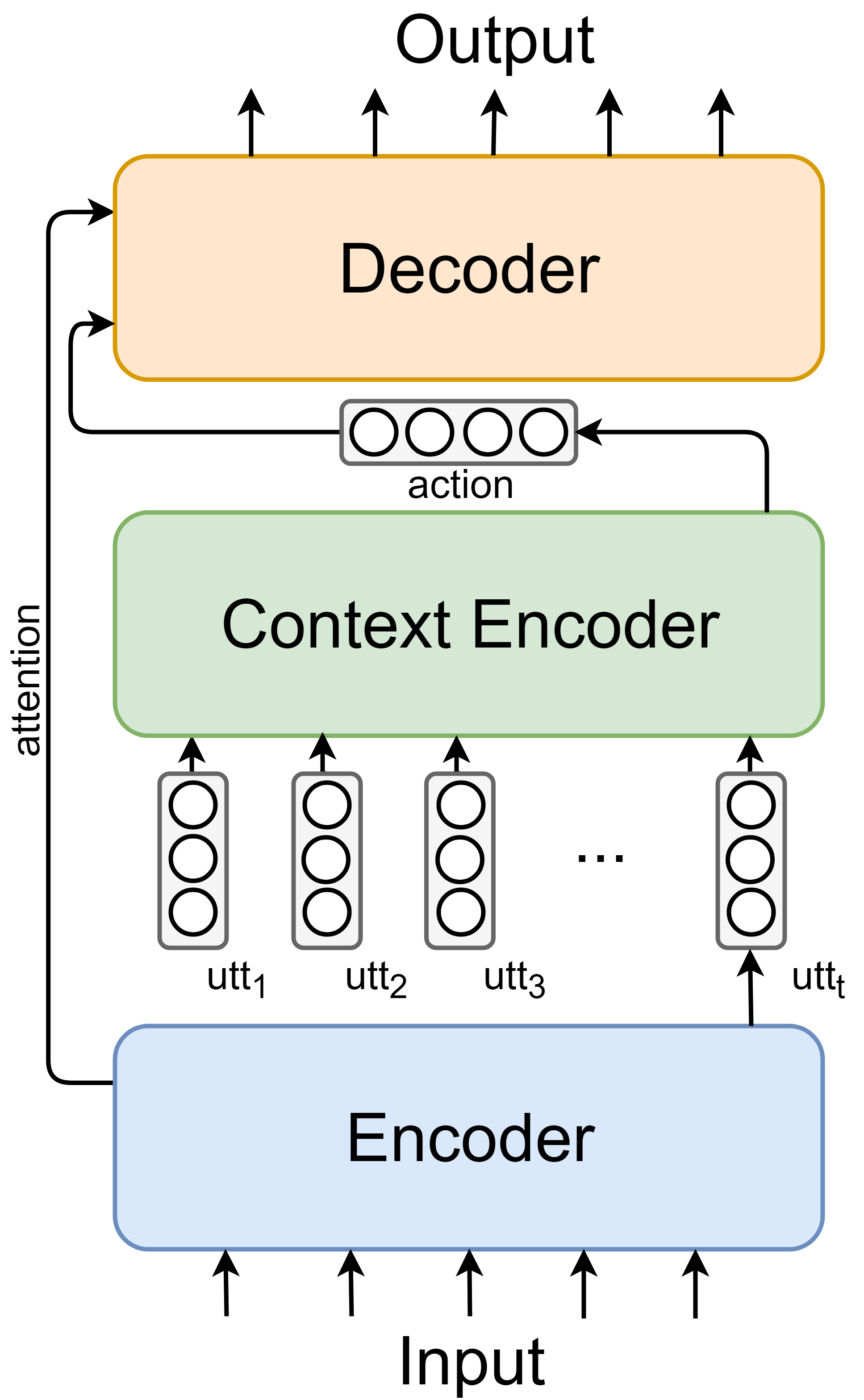}
    \caption{The universal model architecture}
    \label{fig:HRED}
\end{figure}

\subsection{A Universal Dialogue Generation System}
A typical method in dialogue generation is to use a sequence to sequence model\cite{cho2014learning}. The user's input utterance $u_t$ contains a sequence of words $(w^u_{t,0}, w^u_{t,1}, ..., w^u_{t,m})$. The encoder part of a Seq2Seq model takes the words as the input and learning a representation $\mathbf{v}^u_t$ of the utterance $u_t$. And then the action is made from the collection of all utterance representation $(\mathbf{v}^u_{0\sim t})$. 

We use an encoder-decoder model to encode the user utterance to a latent vector representation and summarize the all utterance' vectors with a context-level  encoder in hierarchical encoder-decoder architecture as shown in Figure \ref{fig:HRED}. For an utterance $u_t$ at the time $t$  contains $m$ words $(\mathbf{w}_0, \mathbf{w}_1, ..., \mathbf{w}_m)$. The encoder is an LSTM\cite{hochreiter1997long} network:
\begin{equation}
\mathbf{h}_t=\mathrm{LSTM}_e(\mathbf{h}_0;\mathbf{w}_{t0},\mathbf{w}_{t1},...\mathbf{w}_{tm}),
\end{equation}
Then we consider the last hidden state of the LSTM as the utterance representation vector $\mathbf{v}^u_t=\mathbf{h}_t$, and take the hierarchical encoder as the context-level policy module. The action $\mathbf{a}_t$ is made based on the all history of all utterance. We use another LSTM as the context-level decoder.
$$\mathbf{a}_t=\mathrm{LSTM}_c(\mathbf{v}^u_0, \mathbf{v}^u_1, ..., \mathbf{v}^u_t),$$ The action $\mathbf{a}_t$ is in the form of an abstract latent vector, serving as the guidance for the dialogue system to make proper responses. Although the policy module is not a necessary part in a dialogue system, we'll see how the action representation facilitates the performance of our model using the teacher-student framework.

\begin{figure}[!t]
    \centering
    \includegraphics[width=0.5\linewidth]{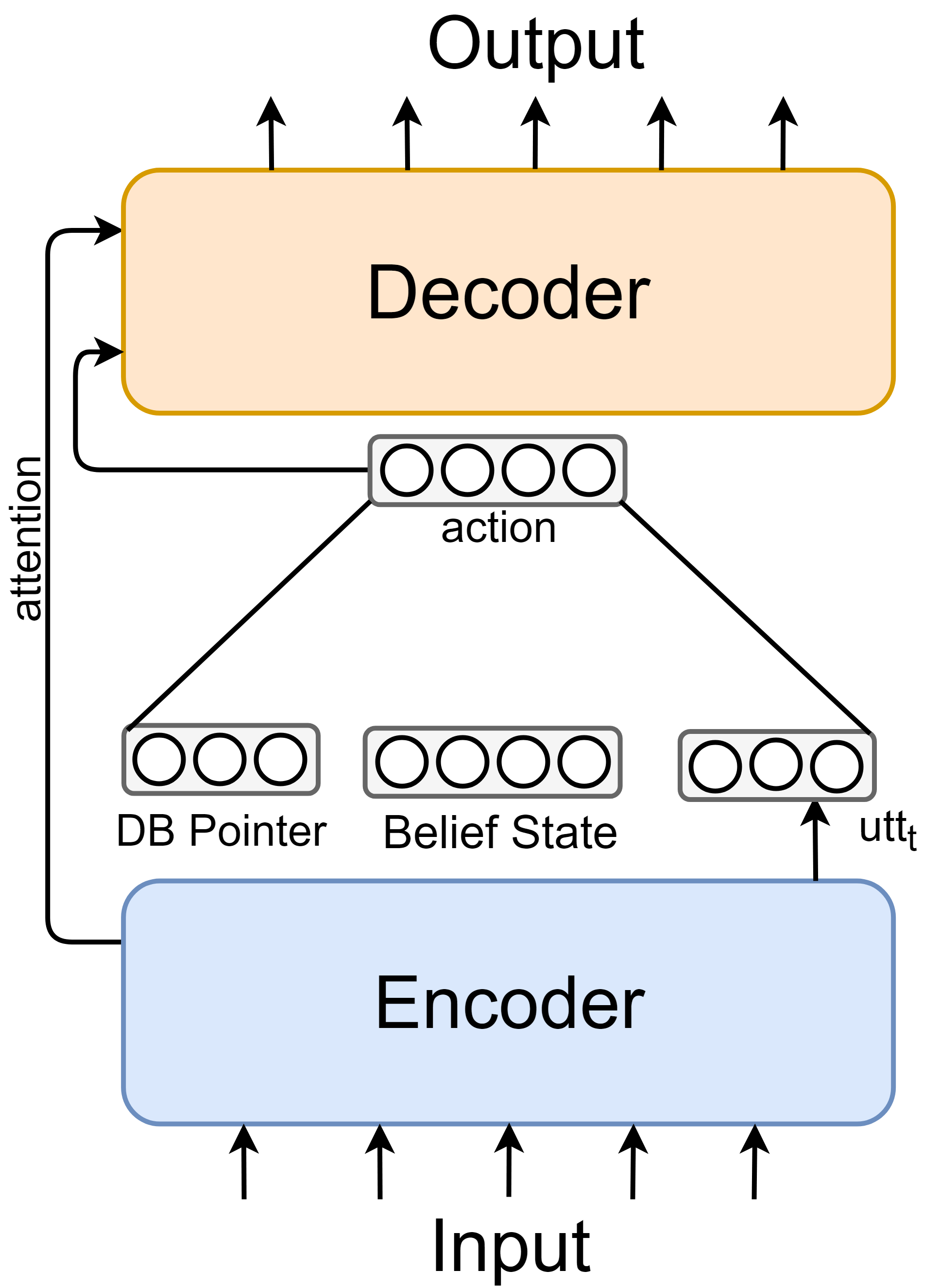}
    \caption{The teacher model pre-trained from each domain}
    \label{fig:baselines}
\end{figure}

\begin{figure*}[ht]
    \centering
    \includegraphics[width=0.8\linewidth]{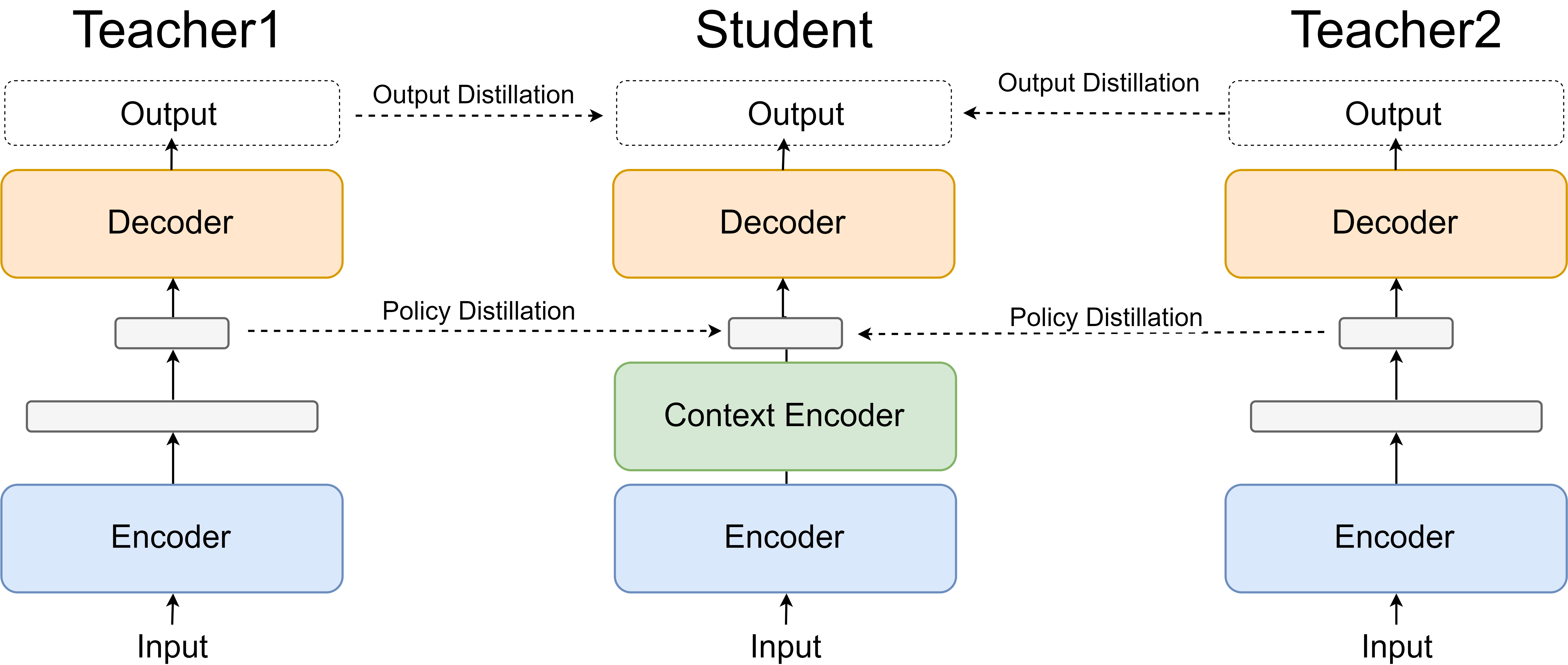}
    \caption{The teacher-student framework that transfers the knowledge from teachers to the student.}
    \label{fig:TS}
\end{figure*}

The action is fed into the generation part lately. The NLG module takes the action as the initial state and generates the final response $r_t$. With the attention mechanism enhanced, the decoder model can be written as:
\begin{equation}
    \mathbf{v}^r_i=\mathrm{LSTM}_d(\mathbf{a}_t,\mathbf{v}^w_{0\sim m},\mathbf{v}^r_{0\sim i-1}),
\end{equation}
where $\mathbf{v}^w_j$ is the output of the encoder in the position of the $j$-th word .

\subsection{Individual Models as Teachers}
We also train dialogue models in each individual domain. Differ from the universal model, the individual models don't generate the actions directly from the utterance. Instead, we use human labeled semantic to construct the states of the conversation.

We use the model proposed by \citet{budzianowski2018towards}. As shown in Figure \ref{fig:baselines}, it contains three parts: the encoder and the decoder as same as the hierarchical model, and a middle policy model that takes both the utterance representation $u_t$ as well as and human defined feature $e_t$ as the input. The feature is split into two vector representations. The first part is the belief state vector $\mathbf{v}_b$, where each dimension of the vector stands for the one-hot value of a specific slot in each domain, a slot that should be received from the user. Thus the whole values of $\mathbf{v}_\mathrm b$ are the necessary information the system keeps at the current situation. At every turn the belief state is updated according to the semantic labeling of the user. Another construction of the state is the database pointer vector $\mathbf{v}_\mathrm{kb}$, where a database pointer vector stands for the number of the corresponding entities that match the request of the user. We use a 6-dimension one-hot embedding vector and each position embedding means separately 0, 1, 2, 3, 4 and more than 4 candidate entities. We concatenate three vectors: the utterance vector $\mathbf{v}^u_t$, the belief state $\mathbf{v}_\mathrm{b}$, and the database pointer $\mathbf{v}_\mathrm{kb}$, to get the vector of the current stage $s_t$ in the conversation.

After the state vector is calculated, we feed the vector to the policy model. The vector is processed with a nonlinear layer with tanh as the activation function, and the action vector $\mathbf{a}_t$ is generated from this layer:
\begin{equation}
\mathbf{a}_t = \mathrm{tanh}(\mathbf{w}\cdot[\mathbf{v}^u_t;\mathbf{v}_\mathrm{b};\mathbf{v}_\mathrm{kb}]),
\end{equation}
where $[;]$ stands for concatenation. The action $\mathbf{a}_t$ is finally delivered to the decoder module and the response is generated with an attention mechanism as mentioned above. 
We train teacher models individually at each domain. Thus the meaning of the belief state differs in teachers. After the teachers are well pre-trained in all domains. We take the teachers as guidance to train the student model.

\section{The Teacher-Student Framework}
\label{t}
This section we described the method to extract the knowledge from individual models in each domain to the universal model. We consider the dialogue problem a mapping from a mixed set $\{\mathcal{X}, \mathcal{C}\}$ to the set $\mathcal{Y}$. $\mathcal{X}$ is the collection of the user's inputs. $\mathcal{C}$ stands for the set of the context or the history during a dialogue. And $\mathcal{Y}$ is the response the system gives. For a well-formed domain-specific dialogue data, the semantic labeling is offered to construct the state representation, and then as the extra inputs. That is, the teachers learn the mapping from $\{\mathcal{X}, \mathcal{S}\}$ to the responses $\mathcal{Y}$. With state set $\mathcal{S}$ based on human labeling, we can take it for granted that teachers learn a better response strategy than the one without that. Then, while the training of the student model, we use extra information from these teachers to guide the training step. We hope to get a final student model that performs as well as the teachers without extra state information. We use the concept of knowledge distillation and guides the student from both the output part and the decision making part to reach as close in performance to the teacher models.
\subsection{Output Guiding}
We first trained our model to learn proper responses from single-domain models. For a generation model at the time given user utterance $u$ and the context $c$, the purpose of the model is to find the most suitable response $r$ with a sequence of words $(w^r_0, w^r_1, ..., w^r_m)$. That can be written as:
\begin{equation}
\begin{split}
    r &= \{w^r_0,w^r_1,...w^r_m\} \\
    &=\mathrm{arg}\max_{w^r\in \mathcal{W}}\prod^m_{i=0}{p(w^r_i|u, c, w^r_{0\sim i-1}; \theta)},
\end{split}
\end{equation}
the $\mathcal{W}$ is the vocabulary of all possible words and $\theta$ is the parameters of the generation model. To apply the guidance from the teachers' output, the student should output a similar result as the teachers do. With a log-likelihood format, the guide method can be written as:
\begin{equation}
\begin{split}
    w^r_i=\mathrm{arg}\max_{w^r\in\mathcal{W}}p(w^r_i|u,s, w^r_{0\sim i-1};\theta_T)\\\log{p(w^r_i|u,c, w^r_{0\sim i-1};\theta)},
\end{split}
\end{equation}
$\theta_T$ is the parameter of the teacher models. Rather than  simply learns the grounding-truth at each turn, the student model also tends to learn the response the teachers give.

We applied 2 methods of distillation:

\textbf{Vocabulary size distillation}. The output logits at each position are totally used for knowledge distillation. This is the naive way to distill the knowledge from the teacher model. For the grounding truth of the training data, the generation part of the model learns only the one-hot value at each position. For the distillation, the guidance from the teachers' output applies a smoother distribution of the probability of words. The vocabulary size distillation brings nature and correctness for the dialogue generation.

\textbf{Top-K distillation}. Only top k logits at each position are used for knowledge distillation\cite{tan2019multilingual}. This method of distillation doesn't use the full probability of the vocabulary size at every step of response generation. The top k of the logits are selected in the teachers' output to be the guidance of the student's training. This kind method of distillation is more efficient than a full vocabulary one. Actually, not all of the words should be taken into consideration in the generation of the sentence so the omitting of the low probability words helps in the guiding process.

% \textbf{Sequence-level distillation}. Also called beam distillation\cite{kim2016sequence}.

We use both the grounding truth and the teachers' output as the target data in the training, and apply the negative of the log-likelihood as the loss of output distillation. The loss is added to the loss of the grounding truth. To adjust the effect of the teachers, we apply a weighted scalar $\alpha_1$ to change the importance of teachers while training.

\subsection{Policy Guiding}
Beside applying distillation in the final outputs, we also expect the universal model learns more parts of the teachers'. The teacher model and the student model differ in the structure of the policy part but have the same decoder module as the NLG part. We can assume that the teacher models and the student model should have similar decision making as input in the NLG model. Thus the teaching process of the policy part from teacher models is helpful to a student model. We use the action $\mathbf a^\text T$ from the teachers' policy output as the extra info to train the policy of student's. For $\mathbf a^\text T$ and $\mathbf a^\text S$ are both in the form of latent vectors. While training we use mean squared error(MSE) loss to force the student to make decisions like the teachers. That is:
\begin{equation}
    J_{\text {KD}-\pi}=\sum^k_{i=0}(a^\text T_i-a^\text S_i)^2,
\end{equation}
While training we add the policy distillation loss $J_{\text {KD}-\pi}$ to the existing loss by multiplying another weighted scalar $\alpha_2$ as the output distillation does.

\section{Experiments}
To figure out the performance of the approaches mentioned above, we apply our model to a multi-domain dialogue problem to test the ability of the teacher-student based dialogue systems.

\begin{table*}[!ht]
\begin{tabular}{c|ccc|ccc}
     \hline
     & \multicolumn{3}{c|}{Multi-domain}
     &  \multicolumn{3}{c}{restaurant}\\
     model & BLEU & Inform(\%) & Success(\%) & BLEU & Inform(\%) & Success(\%) \\
     \hline
     Seq2seq(no states) & \textbf{0.180} & 60.5 & 41.0 & \textbf{0.182} & 79.6 & 68.0 \\
     Seq2seq(GCE states) & 0.162 & 65.2 & 40.6 & 0.163 & 91.9 & 80.0  \\
     HRED & 0.175 & 66.0 & 53.3  & 0.173 & 84.1 & 72.0 \\
     HRED-TS & 0.175 & \textbf{70.0} & \textbf{58.0} & 0.170 & \textbf{92.1} & \textbf{83.4} \\
     \hline
     Seq2seq(Manual states) & 0.189 & 70.6 & 60.0 & 0.191 & 90.1 & 82.3 \\
     \hline
\end{tabular}
\caption{The results between the raw model, the model uses GCE as a state tracker and the model uses knowledge distillation. The Seq2seq(Manual states) model takes the manual labeled states as input and sets as the upper bound of our metrics.}
\label{tbl:table1}
\end{table*}

\subsection{Dataset}
We choose \textbf{MultiWOZ}\cite{budzianowski2018multiwoz}, a multi-domain human-human conversation dataset. The MultiWOZ dataset consists of dialogue data in 7 domains, which vary in restaurant, hotel, attraction, taxi, train, hospital and police. The conversation in the dataset aims at satisfying the intent of the user's, and apply the necessary information the user needs about some entities. An episode of conversation contains around 14 turns of dialogues between the user and the system. Several episodes' topics are limited in one domain from beginning to the end turn, while others' are switching among the conversation in 2 to up 5 domains. In each domain, there are about 4 slots that the system can be received from the user and about 3 properties of the entity the system should provide to the user. For example, in a restaurant domain, the user can choose the area, the price range and the food type of a restaurant, and the information the system should offer about the restaurant includes the address, the reference number, the phone number, etc.

To test the response ability of the models, at first, we take a pre-processing on the dialogue data, replacing the name of the entities and their property values with placeholders. Then we manually generate the belief states and the database pointers, as the extra inputs of teachers, from the human labeled semantics. To train the individual teachers in different domains, we split the dataset into domains in turn-level, tagging the domain of each turn by the entities mentioned in user, system or the human defined dialogue actions. For some episodes may involve more than one domain, One episode may be taken apart into several. Though the fluency of the conversation may be influenced by the lacking of the context in training individual teachers, we think the teacher models can be well trained as they take the turn-level sentences as input as well as the manual state.

\begin{table}[!ht]
\begin{tabular}{c|ccc}
     \hline
     & \multicolumn{3}{c}{Multi-domain} \\
     distill & BLEU & Inform(\%) & Success(\%)  \\
     \hline
     vocab-size &0.175 & 69.3 & 56.6 \\
     top-1 & 0.180& 68.0 & 57.8 \\
     top-8 & 0.172 & 67.4 & 56.0 \\
     top-32 & \textbf{0.184} & 69.2 & 57.3 \\
     top-128  & 0.175 & \textbf{70.0} & \textbf{58.0} \\
     \hline
\end{tabular}
\caption{The result between different output distillation methods.}
\label{tbl:table2}
\end{table}

\subsection{Comparison}
We set several models as the comparison of our models:

\textbf{The universal model without teachers.} We use an HRED model without any teacher's guide. The model directly learns from the raw data, to see how much the teacher-student framework helps in the model training.

\textbf{The dialogue model with the belief state.} We use a dialogue model takes belief state as the input, as well as a state tracking model. The dialogue model is the same as the teacher model in section \ref{t}, takes the belief states as the part of the input. And we also use a Globally Conditioned Encoder(GCE)\cite{budzianowski2018towards} state tracking model, which is the best state tracking model on the MultiWOZ dataset, to update the state as the dialogue carries on. We test this model to see if our framework can successfully summarize the abstract state and make responses from utterance rather than from a belief state.

\textbf{The dialogue model with the manual state.} We also apply a universal dialogue model alone use the manual state as the input during measurement. That is, this model takes more input data than others in comparison. The test result is set as the upper bound as other models can reach. 

\begin{table}[!ht]
\begin{tabular}{c|ccc}
     \hline
     & \multicolumn{3}{c}{Multi-domain} \\
     distill part & BLEU & Inform(\%) & Success(\%)  \\
     \hline
     All & \textbf{0.178} & 67.8 & \textbf{57.0}  \\
     output only & 0.175 & \textbf{69.3} & 56.6 \\
     policy only & 0.174 & 67.1 & 55.4 \\
     universal & 0.176 & 67.1 & 56.8 \\
     \hline
\end{tabular}
\caption{The results between adding different components in a distillation step. The last column shows the results of distilling from a universal teacher model.}
\label{tbl:table3}
\end{table}

\subsection{Settings and metrics}
We use 50 dimensions of the vectors in the word embedding. The vocabulary size is limited to 400 for both the input and the output vocabulary size separately. The HRED model has three parts of LSTM architects, with their hidden size 150. The teacher models have the encoder and the decoder part of 150 dimension LSTM networks as well. For each teacher, we trained it on its individual domain, and find the model has the best performance as the guidance. For the student model, we use Adam optimizer and the learning rate is 0.005. To balance the data training and the teacher guidance, the output distillation loss has a weight $\alpha_1=0.01$ and the weight for policy one $\alpha_2$ is set to 0.05.

To measure the performance of different models, we use several metrics to judge the responses generated. Firstly, we calculate sentence level BLEU-4 scores to measure the similarity of the real response and the generated one. For BLEU scores show less correlation about the quality of the dialogue content, we apply other measurements. We use two metrics that are suggested by \citet{budzianowski2018multiwoz}, as the estimations for the MultiWOZ dataset in the dialogue context to text task. Both the measurements are on the episode level. The \textbf{Inform rate} indicates whether the dialogue system suggests suitable entities according to the user's intent in an episode. And the \textbf{Success rate} illustrates if the system provides all the correct properties the user requests after a success informing. We run the models on the test dataset which includes 1000 episodes of conversations, then count out the ratios of the successfully informed dialogues and totally succeed ones.

\section{Result \& Analysis}
The comparison between the different models is shown in Table \ref{tbl:table1}. From the table, we can see that in multi-domain our model(HRED-TS) gets the best scores in both the informing rate and the success rate. By adding a teacher-student framework, the informing rate and the success rate receive 4\% improvement than the original model. Our model reaches a result as close as the one that takes manual states as input, which is considered as an upper bound. The model with a GCE state tracker increases the Informing rate but has no effect on improving the dialogue success rate. We think that thought the state based on a state tracker has its pros on representing the dialogue processing, the errors exist in the state tracker harms the performance of the model. And our model avoids such a problem by using the teacher-student framework.

In Table \ref{tbl:table1} we also measure our dialogue models' performance in the biggest single domain, the restaurant domain. Results show that our model produces the best results in all models and even outperforms than the one with manual state input. We trust that it is due to using an individual teacher in the restaurant domain while training, which results in a better performance in this domain than the universal one. It is worth noticing that a GCE state has a higher informing rate than the manual state. We believe the lower informing score in manual state one is caused by the influence of state labeling from other domains. 

Table \ref{tbl:table2} shows the effect of the top-k distillation method for our framework, it is clear that the total success rate increases by applying the top-k distillation except the $k$ value is set to 8. Top-128 brings the best performance for the model. It is well explainable that the top-k distillation removes the unnecessary information the teachers give to the student. Meanwhile, the significant words are kept in the distillation to ensure the response is good enough. As the size of the distillation reduced, both the success rate and the informing rate decrease due to the loss of enough guidance. The result shows slightly abnormal when $k=1$. We believe that top-1 distillation is in the same form as to add more grounding truth to the training data.

From Table \ref{tbl:table3}, we can see the results of using different guidance. Adding a policy distillation alone brings the slightest improvement to the original model. The output distillation has the highest informing rate in all guidance methods. By adding both a policy distillation and an output distillation, the model succeeded more often on the dialogue but failed on informing rate, which indicates the decision helps less on entity suggestion and has more effect on properties informing, which involves more decision makings on multiple turns. We also proofed that using a single universal model as the teacher is not well as using individual teachers, for the informing rate of the universal distillation is lower than the output distillation one.

\section{Conclusions}
In the paper, we propose a multi-domain dialogue generation model trained with a teacher-student framework. The model takes only raw text as input and takes full advantage of the human labeled states during training. The model behaves better than the one using an external state tracker, with great improvements in the success rate during a conversation.

The problem exists in our model that it focuses on the text generation during the conversation, and takes no consideration of the knowledge base querying. So our model cannot be regarded as a complete dialogue system. But we don't think it is unable to process. Adding an extra component such as a memory network can solve the problem.

\bibliography{emnlp-ijcnlp-2019}
\bibliographystyle{acl_natbib}

\end{document}